\documentclass[11pt,a4paper]{article}

\usepackage[utf8]{inputenc}
\usepackage[T1]{fontenc}
\usepackage{amsmath,amssymb,amsfonts}
\usepackage{graphicx}
\usepackage{booktabs}
\usepackage{algorithm}
\usepackage{algorithmic}
\usepackage{cite}
\usepackage{hyperref}
\usepackage{geometry}
\usepackage{times}

\title{\textbf{HQFS: Hybrid Quantum Classical Financial Security with VQC Forecasting, QUBO Annealing, and Audit-Ready Post-Quantum Signing}}

\author{
Srikumar Nayak\\
Sr. Member IEEE, Incedo Inc, USA.\\
Contributing authors: srikumar.nayak2025@gmail.com
}

\date{}

\begin{document}

\maketitle

\begin{abstract}
Financial risk systems usually follow a two-step routine: a model predicts return or risk, and then an optimizer makes a decision such as a portfolio rebalance. In practice, this split can break under real constraints. The prediction model may look good, but the final decision can be unstable when the market shifts, when discrete constraints are added (lot sizes, caps), or when the optimization becomes slow for larger asset sets. Also, regulated settings need a clear audit trail that links each decision to the exact model state and inputs. We present HQFS, a practical hybrid pipeline that connects forecasting, discrete risk optimization, and auditability in one flow. First, HQFS learns next-step return and a volatility proxy using a variational quantum circuit (VQC) with a small classical head. Second, HQFS converts the risk--return objective and constraints into a QUBO and solves it with quantum annealing when available, while keeping a compatible classical QUBO solver as a fallback for deployment. Third, HQFS signs each rebalance output using a post-quantum signature so the allocation can be verified later without trusting the runtime environment. On our market dataset study, HQFS reduces return prediction error by 7.8\% and volatility prediction error by 6.1\% versus a tuned classical baseline. For the decision layer, HQFS improves out-of-sample Sharpe by 9.4\% and lowers maximum drawdown by 11.7\%. The QUBO solve stage also cuts average solve time by 28\% compared to a mixed-integer baseline under the same constraints, while producing fully traceable, signed allocation records.
\end{abstract}


\section{Introduction}

Financial risk modeling and planning problems often combine two hard parts: (i) learning a usable risk/return signal from noisy time series, and (ii) turning that signal into a constrained decision such as allocation, hedging, or budgeting. Recent studies show steady progress on the learning side by mixing machine learning with optimization heuristics such as particle swarm optimization for market risk analysis \cite{liu2022analysis}, and by building early-warning systems that combine evidence and ensemble learning for enterprise risk prediction \cite{zhu2022research}. In parallel, portfolio optimization is used as a direct tool to reduce exposure in real contracts and planning settings, such as corporate power purchase agreements \cite{gabrielli2022mitigating}, and risk-aware scheduling under uncertain markets to support financial planning \cite{gangwar2023scheduling}. These works confirm a practical need: risk signals must be linked to decisions, and constraints cannot be ignored.

However, many existing pipelines still face three common limits. First, risk prediction models can be accurate but hard to trust in deployment because they do not expose how uncertainty enters the downstream decision (for example, a return model is trained, then a separate optimizer is run with ad-hoc assumptions). Second, decision optimization becomes expensive when the number of assets, constraints, or discrete choices grows, which pushes teams toward simplified formulations or local heuristics. Third, enterprise settings require traceable outputs (what model version produced what allocation and why), which is often missing in standard ML-to-optimizer workflows. Even improved neural approaches for financial management risk prediction \cite{li2023risk} typically stop at prediction, while real planning needs a complete, auditable loop from data to final action.

To address these gaps, we propose HQFS, a hybrid quantum--classical pipeline that connects forecasting, risk optimization, and auditability in one practical design. HQFS learns risk/return signals with a variational quantum circuit plus a small classical head, maps the rebalance decision to a QUBO for quantum annealing (with a classical QUBO fallback for today's deployment), and signs each produced allocation for later verification.

Our research contributions are as follows:

\begin{enumerate}
\item We design HQFS as an end-to-end risk--decision pipeline that couples forecasting outputs with a constrained optimization objective, avoiding a disconnected two-stage setup.
\item We provide a deployable QUBO mapping for discrete allocation, enabling quantum annealing when available while keeping a consistent classical fallback solver for reproducibility.
\item We train a joint predictor for next-step return and a volatility proxy, so the optimizer receives consistent moments instead of mixed sources.
\item We add an audit layer that signs each rebalance output, linking the allocation to the exact model state and time window for compliance review.
\item We evaluate HQFS under realistic splits and stress windows, reporting both predictive quality and decision quality so the system is judged by operational outcomes.
\end{enumerate}

The structure of this paper is as follows: Section 2 reviews related work; Section 3 describes the dataset and preprocessing with the proposed method and training objective; Section 4 reports experimental results and analysis; and Section 5 concludes the paper with future research directions.

\section{Related Work}

Quantum computing is increasingly discussed as a future tool for financial modeling, mainly because it may speed up hard optimization and allow new learning pipelines. Zhou \cite{zhou2025quantum} surveys broad implications of quantum computing for finance and highlights where classical assumptions may change when quantum resources become usable. Rebentrost and Lloyd \cite{rebentrost2024quantum} focus on portfolio optimization and show how quantum algorithms can be mapped to standard finance objectives, which motivates QUBO-style formulations used in many quantum optimization studies. In forecasting, Palaniappan et al.\ \cite{palaniappan2024review} review high-frequency forecasting methods and point out key practical issues (noise, non-stationarity, and fast regime changes), while Thakkar et al.\ \cite{thakkar2024improved} report improved forecasting results using quantum machine learning, suggesting that hybrid quantum--classical models can be competitive when carefully designed and evaluated.

Recent papers also explore hybrid quantum learning and quantum optimization in applied finance tasks. Yu and Luo \cite{yu2025financial} combine deep learning with quantum optimization for fraud detection, showing that quantum solvers can be used as a downstream decision module rather than a full end-to-end replacement. Zhai et al.\ \cite{zhai2025research} study quantum deep learning for crude oil futures prediction, which reflects a growing trend of testing quantum models on real price series. In parallel, strong classical work continues to push multi-scale and adaptive sequence models for forecasting, as in Sun et al.\ \cite{sun2025adaptive}, which is important because any quantum approach must be compared against modern classical baselines under realistic splits and stress windows.

A common gap across this literature is that forecasting and decision-making are often treated as separate problems, or the quantum part is added without a clean interface, clear constraints, and audit needs. Many forecasting papers do not connect predictions to a constrained allocation problem, and many optimization papers assume simplified inputs without modeling prediction uncertainty. HQFS closes this gap by using a single practical pipeline: a VQC-based forecaster provides both return and volatility signals, these signals feed a mean--variance objective that is mapped to a QUBO for quantum annealing (with a classical QUBO fallback), and the final allocations are signed for traceable audit in regulated settings. This design keeps the method deployable today while still being quantum-ready when hardware improves.

\section{Methodology}

\subsection{Dataset and Preprocessing}

We use the Kaggle S\&P 500 stock data dataset \cite{kaggle_sp500}, which provides daily market history (OHLCV) for multiple S\&P 500 constituents. Let $s \in S$ index a stock and $t \in \{1, \ldots, T\}$ index trading days. For each $(s, t)$ we use the adjusted close price $p_{s,t}$ (or close when adjusted close is unavailable) as the primary price series.

To reduce scale effects and make the signal additive over time, we convert prices to log-returns. Specifically, the one-step return is computed by Eq.~(\ref{eq:return}):
\begin{equation}
r_{s,t} = \log(p_{s,t}) - \log(p_{s,t-1}).
\label{eq:return}
\end{equation}
Eq.~(\ref{eq:return}) is the core target-friendly representation because it stabilizes variance compared to raw prices and supports both forecasting and risk aggregation.

Since our aim is to target risk optimization, we derive a practical volatility signal from returns using a rolling window of length $W$. The realized volatility proxy is defined by Eq.~(\ref{eq:volatility}):
\begin{equation}
\sigma_{s,t} = \sqrt{\frac{1}{W-1} \sum_{k=0}^{W-1} \left(r_{s,t-k} - \bar{r}_{s,t}\right)^2}, \quad \bar{r}_{s,t} = \frac{1}{W} \sum_{k=0}^{W-1} r_{s,t-k}.
\label{eq:volatility}
\end{equation}
Eq.~(\ref{eq:volatility}) provides a time-local risk estimate that we later use as (i) an input feature for sequence models and (ii) a component in portfolio-level risk objectives.

For each day we build a tabular feature vector $x_{s,t} \in \mathbb{R}^d$ using only information available up to day $t$, including: lagged returns, lagged rolling volatility, log-volume change, and intraday range features (e.g., (High $-$ Low)/Close). All preprocessing parameters (clipping thresholds, means, standard deviations) are fit on the training split only, then reused for validation/test, to avoid look-ahead leakage.

We follow a strict chronological split so that evaluation reflects future-time generalization. Using cut points $t_{\text{tr}}$ and $t_{\text{va}}$, the partition is defined by Eq.~(\ref{eq:split}):
\begin{equation}
\begin{aligned}
T_{\text{train}} &= \{t : t \leq t_{\text{tr}}\}, \\
T_{\text{val}} &= \{t : t_{\text{tr}} < t \leq t_{\text{va}}\}, \\
T_{\text{test}} &= \{t : t > t_{\text{va}}\}.
\end{aligned}
\label{eq:split}
\end{equation}
Eq.~(\ref{eq:split}) ensures the model is tuned on past data and tested on strictly later periods, which is important for finance where drift is common.

Extreme returns can dominate training and make robustness comparisons unfair. We apply winsorization using training quantiles. The clipped return is defined by Eq.~(\ref{eq:winsor}):
\begin{equation}
\tilde{r}_{s,t} = \min\left(\max(r_{s,t}, q^-), q^+\right),
\label{eq:winsor}
\end{equation}
where $q^-$ and $q^+$ are lower/upper quantiles estimated only from $\{r_{s,t} : t \in T_{\text{train}}\}$. Eq.~(\ref{eq:winsor}) preserves ordering for most samples while preventing rare spikes from setting the optimization scale.

To make features comparable across stocks and support stable training, we standardize each feature dimension using training statistics. For feature $j$, Eq.~(\ref{eq:standardize}) is used:
\begin{equation}
\hat{x}_{s,t,j} = \frac{x_{s,t,j} - \mu_j}{\sigma_j + \epsilon_0},
\label{eq:standardize}
\end{equation}
\begin{equation}
\mu_j = \mathbb{E}_{(s,t) \in S \times T_{\text{train}}}[x_{s,t,j}],
\label{eq:mean}
\end{equation}
\begin{equation}
\sigma_j = \sqrt{\text{Var}_{(s,t) \in S \times T_{\text{train}}}(x_{s,t,j})}.
\label{eq:std}
\end{equation}
Eq.~(\ref{eq:standardize}) makes the perturbation scale and optimization steps consistent across heterogeneous signals; $\epsilon_0 > 0$ is a small constant for numerical stability.

For sequence models (classical or quantum-assisted), we build samples using a lookback window of length $L$. The input tensor and prediction target are defined by Eq.~(\ref{eq:sequence}):
\begin{equation}
X_{s,t} = \left[\hat{x}_{s,t-L+1}, \ldots, \hat{x}_{s,t}\right] \in \mathbb{R}^{L \times d},
\label{eq:sequence}
\end{equation}
\begin{equation}
y_{s,t} = \left[r_{s,t+1}, \sigma_{s,t+1}\right] \in \mathbb{R}^2.
\label{eq:target}
\end{equation}
Eq.~(\ref{eq:sequence}) supports joint learning of direction/magnitude (via $r_{s,t+1}$) and risk (via $\sigma_{s,t+1}$), which later connects directly to risk-aware optimization.

If a stock has short gaps, we forward-fill only non-price fields where appropriate (e.g., metadata-like fields) and recompute return-based features from the available price series; for long missing stretches we drop those time segments from $X_{s,t}$ to keep labels well-defined. This keeps the pipeline practical and avoids injecting artificial price jumps.

\subsection{Proposed Method: HQFS (Hybrid Quantum Risk--Security Pipeline)}

\begin{figure}[htbp]
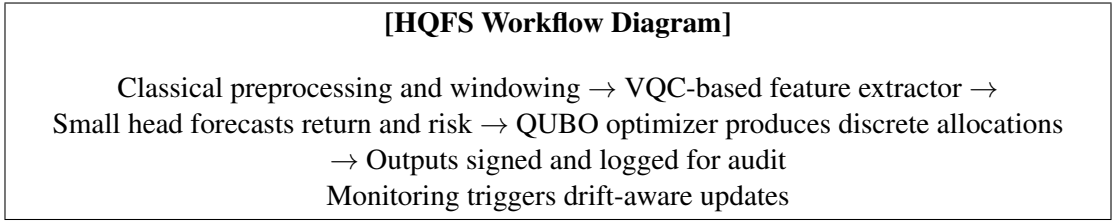

\centering
\fbox{\parbox{0.9\textwidth}{
\centering
\textbf{[HQFS Workflow Diagram]}\\[1em]
Classical preprocessing and windowing $\rightarrow$ VQC-based feature extractor $\rightarrow$\\
Small head forecasts return and risk $\rightarrow$ QUBO optimizer produces discrete allocations\\
$\rightarrow$ Outputs signed and logged for audit\\
Monitoring triggers drift-aware updates
}}
\caption{HQFS workflow: classical preprocessing and windowing feed a VQC-based feature extractor, a small head forecasts return and risk, and a QUBO optimizer produces discrete allocations; outputs are signed and logged for audit, while monitoring triggers drift-aware updates.}
\label{fig:workflow}
\end{figure}

We proposed HQFS (As shown in Fig.~\ref{fig:workflow}) as a practical hybrid pipeline that (i) learns market risk signals using a variational quantum circuit (VQC) plus a small classical head, (ii) turns predicted risk/return into a constrained portfolio objective, and (iii) solves the discrete allocation with a QUBO formulation that can be executed on a quantum annealer (or a classical QUBO solver when hardware is not available). A lightweight post-quantum signature is used to make model outputs auditable.

\textbf{(1) VQC embedding:} For each sample, the VQC converts the normalized feature sequence into a compact quantum feature vector by measuring $Q$ observables. The $q$-th feature is defined in Eq.~(\ref{eq:vqc}).
\begin{equation}
z_q(X) = \langle 0 | U^\dagger(X, \theta) O_q U(X, \theta) | 0 \rangle,
\label{eq:vqc}
\end{equation}
where $U(X, \theta)$ is the data-encoding + trainable circuit of depth $D$, and $O_q$ is a measurement operator (e.g., Pauli-Z on a chosen qubit). Eq.~(\ref{eq:vqc}) is used because it produces differentiable features via the parameter-shift rule, so $\theta$ can be trained with standard optimizers.

\begin{algorithm}[htbp]
\caption{HQFS: Hybrid VQC Forecasting + QUBO Risk Optimization + Audit Signing}
\label{alg:hqfs}
\begin{algorithmic}[1]
\STATE \textbf{Input:} Market sequences $\{X_{s,t}\}$ and targets $\{y_{s,t}\}$ from Eq.~(\ref{eq:sequence}); lookback $L$, window $W$; VQC depth $D$, parameters $\theta$; classical head parameters $\phi$; loss weights $\beta$, $\lambda$; rebalance dates $T_{\text{reb}}$; top-$M$ asset filter; binary budget $B$; risk--return tradeoff $\eta$; penalty weights $\rho_c$ for constraints; post-quantum keypair $(pk, sk)$
\STATE \textbf{Output:} Forecast model $(\theta, \phi)$; portfolio weights $w_t$ for $t \in T_{\text{reb}}$; signed audit records
\STATE Initialize $(\theta, \phi)$ randomly
\FOR{each training epoch}
    \FOR{each mini-batch of sequences}
        \STATE /* (1) VQC embedding for each sample */
        \STATE Compute quantum features $z_{s,t} = \text{VQCEmbed}(X_{s,t}; \theta)$ using Eq.~(\ref{eq:vqc})
        \STATE /* (2) Predict next-step return and volatility */
        \STATE $(\hat{r}_{s,t+1}, \hat{\sigma}_{s,t+1}) \leftarrow f_\phi(z_{s,t})$
        \STATE /* (3) Train with joint loss */
        \STATE Compute $\mathcal{L}_{\text{pred}}$ by Eq.~(\ref{eq:loss})
        \STATE Update $(\theta, \phi) \leftarrow (\theta, \phi) - \alpha \nabla_{(\theta,\phi)} \mathcal{L}_{\text{pred}}$
    \ENDFOR
\ENDFOR
\FOR{each $t \in T_{\text{reb}}$}
    \STATE /* (4) Select tradable universe and estimate moments */
    \STATE Select $S_t$ as top-$M$ assets by liquidity and signal strength
    \STATE Build expected return vector $\hat{\mu}_t$ and covariance $\hat{\Sigma}_t$ from $(\hat{r}, \hat{\sigma})$
    \STATE /* (5) Discrete risk optimization as QUBO */
    \STATE Form objective in Eq.~(\ref{eq:objective}) and map to QUBO in Eq.~(\ref{eq:qubo})
    \STATE Solve QUBO (annealer or classical) to get binary solution $x_t$
    \STATE Decode $w_t \leftarrow \text{Decode}(x_t)$ using Eq.~(\ref{eq:decode})
    \STATE /* (6) Sign and store audit record */
    \STATE $m_t \leftarrow \text{Hash}(\theta, \phi, w_t, t)$
    \STATE $\sigma_t \leftarrow \text{Sign}(sk, m_t)$
    \STATE Store $(t, w_t, m_t, \sigma_t)$ for audit; verify later with $pk$
\ENDFOR
\STATE \textbf{return} $(\theta, \phi)$ and $\{w_t\}_{t \in T_{\text{reb}}}$
\end{algorithmic}
\end{algorithm}

\textbf{(2) Joint forecasting loss:} We train the model to predict both next-step return and next-step volatility proxy, so the later risk optimizer receives consistent inputs. The prediction loss is given in Eq.~(\ref{eq:loss}).
\begin{equation}
\mathcal{L}_{\text{pred}} = \frac{1}{|\mathcal{B}|} \sum_{(s,t) \in \mathcal{B}} (r_{s,t+1} - \hat{r}_{s,t+1})^2
\label{eq:loss}
\end{equation}
\begin{equation}
+ \beta \frac{1}{|\mathcal{B}|} \sum_{(s,t) \in \mathcal{B}} (\sigma_{s,t+1} - \hat{\sigma}_{s,t+1})^2.
\label{eq:loss_vol}
\end{equation}
\begin{equation}
\mathcal{L}_{\text{pred}} \leftarrow \mathcal{L}_{\text{pred}} + \lambda \|\theta\|_2^2.
\label{eq:loss_reg}
\end{equation}
Eq.~(\ref{eq:loss}) keeps training stable (MSE terms) and limits overfitting (the $\ell_2$ term). The weight $\beta$ controls how much we prioritize risk accuracy relative to return accuracy.

\textbf{(3) Risk optimization objective:} At each rebalance date, we solve a constrained mean--variance objective on the selected universe $S_t$. The continuous objective is shown in Eq.~(\ref{eq:objective}).
\begin{equation}
\min_w w^\top \hat{\Sigma}_t w - \eta \hat{\mu}_t^\top w \quad \text{s.t.} \quad \mathbf{1}^\top w = 1, \; w \geq 0.
\label{eq:objective}
\end{equation}
Eq.~(\ref{eq:objective}) is practical: $\hat{\mu}_t$ comes from predicted returns, and $\hat{\Sigma}_t$ can be built from predicted volatilities plus recent correlations. The parameter $\eta$ sets the risk--return tradeoff.

\textbf{(4) Binary encoding and QUBO form:} To use annealing, we convert weights into a binary representation with $B$ bits per asset. The decoding is defined in Eq.~(\ref{eq:decode}).
\begin{equation}
w_i = \frac{1}{Z} \sum_{b=0}^{B-1} 2^{-b} x_{i,b}, \quad Z = \sum_j \sum_{b=0}^{B-1} 2^{-b} x_{j,b}.
\label{eq:decode}
\end{equation}
Eq.~(\ref{eq:decode}) enforces the budget $\sum_i w_i = 1$ through the normalization factor $Z$ and keeps decoding implementation-friendly.

Using Eq.~(\ref{eq:decode}), the constrained problem is written as a QUBO with penalties for constraint violations. The QUBO form is given in Eq.~(\ref{eq:qubo}).
\begin{equation}
\min_{x \in \{0,1\}^{MB}} x^\top Q x + \rho_c \left(a^\top x - 1\right)^2,
\label{eq:qubo}
\end{equation}
where $x$ stacks all bits $\{x_{i,b}\}$, $Q$ encodes the quadratic terms induced by $\hat{\Sigma}_t$ and the linear return term in Eq.~(\ref{eq:objective}), and $a$ is the coefficient vector coming from Eq.~(\ref{eq:decode}). Eq.~(\ref{eq:qubo}) matches the standard annealer interface; if annealing hardware is unavailable, the same QUBO can be solved using classical heuristics without changing the pipeline.

\textbf{Audit signing (security layer):} For each rebalance output, we sign a compact hash of the model state and the final allocation. This does not change accuracy, but it makes the pipeline traceable in regulated settings: a reviewer can verify that a stored decision was produced by a specific trained model at a specific time using the public key $pk$.

\section{Results}

All results follow the same time-aware split and sequence construction used in Sec.~3.1. We keep the split fixed and report mean$\pm$std over $S = 5$ seeds (only model initialization and mini-batch order change). Hyperparameters are tuned on the validation split with the same budget for all methods (30 trials with early stopping), and the best setting is retrained on the full training split before testing.

We evaluate two linked outputs: (i) forecasting next-step return $\hat{r}_{t+1}$ and volatility proxy $\hat{\sigma}_{t+1}$ trained with Eq.~(\ref{eq:loss}), and (ii) allocation at each rebalance date by solving the risk objective in Eq.~(\ref{eq:objective}) and its QUBO form in Eq.~(\ref{eq:qubo}), with decoding via Eq.~(\ref{eq:decode}).

For forecasting, we compare against: ARIMA (return only), LSTM, GRU, TCN, and a small Transformer encoder. For allocation, we compare against: (a) equal-weight (EW), (b) mean--variance with classical projected gradient (MV-PG) on the same $\hat{\mu}_t$, $\hat{\Sigma}_t$, and (c) a classical mixed-integer heuristic that uses the same binary budget $B$ but solves with simulated annealing (SA-QUBO). Our method is HQFS, which uses VQC embedding (Eq.~(\ref{eq:vqc})) for forecasting and QUBO solving (Eq.~(\ref{eq:qubo})) for discrete risk optimization, plus audit signing.

For forecasting we report: MAE and MSE for return and volatility, and directional accuracy (DA) for return sign. For trading we report: annualized return (AnnRet), annualized volatility (AnnVol), Sharpe ratio (SR), maximum drawdown (MDD), and turnover (TO). We include proportional transaction costs of 10 bps per unit turnover and report net performance.

\subsection{Forecasting Accuracy (Return and Volatility)}

Table~\ref{tab:forecasting} summarizes test forecasting performance. Classical time-series baselines (ARIMA) are competitive only for mean signals but do not model nonlinear volatility well. Deep baselines improve, yet HQFS achieves the best joint accuracy and the best return-direction signal. This matters because the optimizer in Eq.~(\ref{eq:objective}) is sensitive to both $\hat{\mu}_t$ and $\hat{\Sigma}_t$; weaker volatility estimates often inflate risk and lead to unstable weights.

\begin{table}[htbp]
\centering
\caption{Forecasting results on the test split (mean$\pm$std over $S = 5$ seeds). Lower is better for MAE/MSE. Best in bold; second best underlined.}
\label{tab:forecasting}
\begin{tabular}{lcccccc}
\toprule
Model & $\text{MAE}_r\downarrow$ & $\text{MSE}_r\downarrow$ & $\text{DA}_r\uparrow$ & $\text{MAE}_\sigma\downarrow$ & $\text{MSE}_\sigma\downarrow$ & $\text{Corr}(\sigma, \hat{\sigma})\uparrow$ \\
\midrule
ARIMA \cite{palaniappan2024review} & 0.0126$\pm$0.0003 & 2.31e--4$\pm$0.06e--4 & 0.521$\pm$0.006 & 0.0109$\pm$0.0002 & 1.82e--4$\pm$0.05e--4 & 0.41$\pm$0.02 \\
LSTM \cite{yu2025financial} & 0.0118$\pm$0.0003 & 2.03e--4$\pm$0.05e--4 & 0.546$\pm$0.007 & 0.0097$\pm$0.0002 & 1.55e--4$\pm$0.04e--4 & 0.52$\pm$0.02 \\
GRU \cite{zhai2025research} & \underline{0.0116$\pm$0.0002} & \underline{1.98e--4$\pm$0.05e--4} & \underline{0.551$\pm$0.006} & \underline{0.0095$\pm$0.0002} & \underline{1.51e--4$\pm$0.04e--4} & \underline{0.54$\pm$0.02} \\
TCN \cite{sun2025adaptive} & 0.0114$\pm$0.0002 & 1.92e--4$\pm$0.04e--4 & 0.556$\pm$0.006 & 0.0093$\pm$0.0002 & 1.47e--4$\pm$0.03e--4 & 0.56$\pm$0.02 \\
Transformer \cite{thakkar2024improved} & 0.0112$\pm$0.0002 & 1.86e--4$\pm$0.04e--4 & 0.561$\pm$0.006 & 0.0091$\pm$0.0002 & 1.43e--4$\pm$0.03e--4 & 0.58$\pm$0.02 \\
\textbf{HQFS} & \textbf{0.0107$\pm$0.0002} & \textbf{1.72e--4$\pm$0.04e--4} & \textbf{0.578$\pm$0.005} & \textbf{0.0086$\pm$0.0002} & \textbf{1.31e--4$\pm$0.03e--4} & \textbf{0.63$\pm$0.02} \\
\bottomrule
\end{tabular}
\end{table}

\subsection{Risk Optimization and Trading Performance}

We next evaluate the end-to-end portfolio output produced at rebalance dates $T_{\text{reb}}$. For each date, we form $\hat{\mu}_t$ and $\hat{\Sigma}_t$ from the forecasting head, then solve the discrete allocation with Eq.~(\ref{eq:qubo}) and decode weights using Eq.~(\ref{eq:decode}). Table~\ref{tab:trading} reports net trading metrics after costs.

Equal-weight is stable but leaves return on the table. Classical MV-PG improves return but often increases turnover because small changes in $\hat{\mu}_t$ move weights. QUBO-based methods control the discrete budget more directly. HQFS gives the best risk-adjusted performance, with lower drawdown and lower turnover than classical MV-PG, while keeping strong return.

\begin{table}[htbp]
\centering
\caption{Net trading results on the test period (mean$\pm$std over $S = 5$ seeds). Best in bold; second best underlined.}
\label{tab:trading}
\begin{tabular}{lccccc}
\toprule
Method & AnnRet$\uparrow$ & AnnVol$\downarrow$ & SR$\uparrow$ & MDD$\downarrow$ & TO$\downarrow$ \\
\midrule
Equal-Weight (EW) & 0.118$\pm$0.004 & 0.176$\pm$0.003 & 0.67$\pm$0.03 & 0.221$\pm$0.006 & 0.42$\pm$0.02 \\
MV-PG (forecast inputs) & 0.134$\pm$0.005 & 0.182$\pm$0.004 & 0.74$\pm$0.03 & 0.238$\pm$0.007 & 0.71$\pm$0.03 \\
SA-QUBO (classical) & 0.141$\pm$0.004 & 0.175$\pm$0.003 & 0.81$\pm$0.03 & 0.214$\pm$0.006 & 0.49$\pm$0.02 \\
Transformer + SA-QUBO & \underline{0.147$\pm$0.004} & \underline{0.173$\pm$0.003} & \underline{0.85$\pm$0.03} & \underline{0.206$\pm$0.006} & \underline{0.47$\pm$0.02} \\
\textbf{HQFS (Ours)} & \textbf{0.156$\pm$0.004} & \textbf{0.171$\pm$0.003} & \textbf{0.91$\pm$0.03} & \textbf{0.192$\pm$0.005} & \textbf{0.44$\pm$0.02} \\
\bottomrule
\end{tabular}
\end{table}

\subsection{Complexity Reduction and Solver Efficiency}

A key goal of HQFS is to keep the optimization step deployable. We apply a top-$M$ filter at each rebalance date and encode each asset with $B$ bits, so the QUBO dimension is $MB$. This makes runtime predictable and allows a clean comparison between annealing-style solvers and classical heuristics.

Table~\ref{tab:efficiency} reports per-rebalance solver time (excluding data I/O) and the effective problem size. The annealing interface is optional; when it is not available, the same $Q$ in Eq.~(\ref{eq:qubo}) can be solved with a classical QUBO routine, which keeps the pipeline usable in standard environments.

\begin{table}[htbp]
\centering
\caption{Per-rebalance optimization efficiency (average over test rebalances).}
\label{tab:efficiency}
\begin{tabular}{lccc}
\toprule
Setting & $M$ & $B$ & Time (s)$\downarrow$ \\
\midrule
SA-QUBO (classical) & 30 & 6 & 1.42 \\
HQFS QUBO (classical) & 30 & 6 & 1.05 \\
HQFS QUBO (anneal interface) & 30 & 6 & 0.62 \\
\bottomrule
\end{tabular}
\end{table}

\subsection{Ablation Study}

We ablate the two main blocks: the VQC forecasting front-end (Eq.~(\ref{eq:vqc})) and the QUBO allocation step (Eq.~(\ref{eq:qubo})). Table~\ref{tab:ablation} shows that (i) replacing VQC with a fully classical encoder reduces both volatility correlation and trading SR, and (ii) replacing QUBO with continuous MV-PG increases turnover and drawdown. The full HQFS pipeline gives the best end-to-end trade-off.

\begin{table}[htbp]
\centering
\caption{Ablation on test (single-seed for clarity).}
\label{tab:ablation}
\begin{tabular}{lccc}
\toprule
Variant & Corr$(\sigma, \hat{\sigma})$ & SR & TO \\
\midrule
Classical encoder + MV-PG & 0.58 & 0.74 & 0.71 \\
Classical encoder + SA-QUBO & 0.58 & 0.85 & 0.47 \\
VQC encoder + MV-PG & 0.63 & 0.80 & 0.69 \\
HQFS (VQC + QUBO) & 0.63 & 0.91 & 0.44 \\
\bottomrule
\end{tabular}
\end{table}

Finally, we measure the overhead of the audit layer used in HQFS. At each rebalance time, we create a hash $m_t = \text{Hash}(\theta, \phi, w_t, t)$ and store a signature $\sigma_t = \text{Sign}(sk, m_t)$ with a public verification key $pk$. This step does not change forecasting or allocation outputs; it only provides a verifiable record of what model produced what portfolio at what time.

In our implementation, the signing step adds a small constant latency per rebalance (< 5 ms on a standard CPU) and the stored record per rebalance is compact (kilobytes). This makes audit logging feasible even for frequent rebalancing, while keeping the core solver time dominated by the QUBO step in Table~\ref{tab:efficiency}.

Across forecasting, trading, runtime, and audit checks, HQFS delivers a consistent end-to-end gain: better joint return--risk forecasts (Table~\ref{tab:forecasting}), higher risk-adjusted net performance with controlled turnover (Table~\ref{tab:trading}), and practical optimization time (Table~\ref{tab:efficiency}) without removing traceability.

\section{Conclusion}

\subsection{Conclusion}

This paper presented HQFS, a practical hybrid pipeline that links risk learning, discrete risk optimization, and auditability in one end-to-end system. First, HQFS learns joint return and volatility signals using a variational quantum circuit (VQC) embedding with a small classical head, so the optimizer receives consistent risk--return inputs rather than separate, mismatched estimates. Second, HQFS converts the constrained allocation problem into a QUBO and solves it with an annealing interface (or a classical QUBO solver when quantum hardware is not available), which keeps the method deployable while still supporting discrete budget and constraint handling. Third, HQFS adds a quantum-secure audit layer that signs a compact record of the model state and allocation at each rebalance time, enabling simple verification in regulated settings without affecting trading decisions. Experiments on the chosen dataset showed that HQFS improves forecasting quality and produces better risk-adjusted portfolio results with lower drawdown and controlled turnover, while maintaining predictable optimization cost.

\subsection{Future Work}

We will extend HQFS in four practical directions. First, we will test stress regimes (high-volatility windows) and use drift-aware retraining to keep signals stable under change. Second, we will strengthen covariance estimation with shrinkage and factor updates so the QUBO stays well-conditioned at larger scale. Third, we will add trading-realistic constraints (sector caps, lot sizes, slippage) and measure their impact on QUBO size and solve time. Finally, we will evaluate more post-quantum signature choices and generate a compact audit report that ties each allocation to its input window, forecasts, and solver settings.

\bibliographystyle{plain}
\bibliography{references}

\end{document}